\documentclass[10pt,runningheads]{llncs}
\usepackage{graphicx}
\usepackage{amsmath}
\usepackage{booktabs}
\begin{document}
\title{Question and Answer Classification with Deep Contextualized Transformer}

%
%
\author{Haozheng Luo\inst{1,2}\and
Ningwei Liu\inst{2} \and
Charles Feng\inst{2}}
\authorrunning{}
%
\institute{Georgia Institute of Technology , Atlanta, GA, 30308, USA \and
Splunk, Inc}
\maketitle              
\begin{abstract}
Recent literature has focused on the Standford Parse Tree and how it has been used for Question and Answer problems in Natural Language Processing. This parser tree with deep learning algorithms has  analyzed the makeup of question and answer classifications. In this study, the authors have built a model using a Deep Contextualized Transformer that can manage some aberrant expressions. We conducted extensive evaluations on  SQuAD and SwDA datasets and our model showed significant improvements to QA problem classifications for industry needs. Further analysis investigated the impact of various models on the accuracy of the results. Research outcomes showed that our new method is more effective in solving QA problems with a higher accuracy of up to 83.1\% compared to other models. 

\keywords{QA Classification \and NLP, Self-attention \and Self-attention}
\end{abstract}
\section{Introduction}
The Question and Answer system (QA) is widely used in the industry. Every week, one company faces thousands of questionnaires for the products they launch. QA is a massive problem in Natural Language Processing (NLP), including the application of problem answering, sentence recognitions, etc. There are several types of problems, such as wh-questions, statement questions, statements and other question patterns. Each type of question has a corresponding label for a question or statement. In this study, we want to discover  a better algorithm to analyze the question and answer data from the huge text files. 
Earlier work in this field mainly used the Bag-of-words (BoW) technique to classify sentence types \cite{yogatama2014making}. Many recent studies have adopted supervised, deep-learning methods concerning question and answer classification and have shown promising results \cite{lee2016sequential}.   However, most of these approaches have treated the sentence as a text classification without considering the context of the writing  across sentences or interdependently; therefore, this method  is unable to reflect conceptual dependencies of the words within the sentences. In reality, the different order of the same words in a sentence can have very different meanings. As a result, it is necessary to determine a high compatibility algorithm to classify question and answer sentences by considering all sentence configurations.

This  research draws on some recent advances in NLP research such as BERT \cite{devlin2018bert} and Elmo  \cite{peters2018deep} to produce a sentence classification model to quickly and correctly pick out the question and answer sentence from the target text. Compared with regular algorithms for treating QA problems such as word2vec \cite{pennington2014glove}, this self-attention algorithm can perform contextualized word representation to obtain contextualized word meaning from the sentences. As a result, with BERT and Elmo, we can quickly judge the contexture relationship of the sentence and figure out what kind  sentence type it might be. Specifically, we have used a hierarchical deep neural network with a self-attention algorithm to process different types of question and answer text, including statement questions that are a specific type of question in the questionnaires. This research aims to achieve state-of-the-art outcomes for classifying the QA problem. As a result, we mainly contribute to: 1.)  a huge the improvement of performance on QA classification problems with the self-attention method, especially one such as BERT;   2. demonstrating how performance could be improved with a combination of different levels of models including the hierarchical deep neural network for classification, the self-attention model like BERT for single word embedding, and the previous label of the training data with the SQuAD dataset; and 3. exploring different methods to find a high compatibility method to classify the QA problem.

\section{Related Work}
We focus on four primary methods used in recent research. One treats text as text classification, in which each word is classified in isolation, while the second treats the text using Contextualized Word Representation Algorithms, such as BERT with self-attention or Elmo.  The third method uses the transformer-based foundation models to do the question and answering, and the forth method use the chain-of-thought to do the classfication.

\subsection{Text Classification: }
Lee and Dernoncourt \cite{lee2016sequential} build a vector representing each utterance and use either RNN or CNN to predict the text details to classify the sentence type. 

\subsection{Self-attention}
Jacob et al. \cite{devlin2018bert} used the BERT, and Peters et al. \cite{peters2018deep} used Elmo to embed the text into the vector to give the contextual relationship of the sentence for each word. Along with these two tools, we use RNN-based or CNN-based hierarchical neural networks to learn and model multiple levels of word. 

\subsection{Foundation Model}
Additionally, the development of the transformer architecture has made transformer-based foundation models \cite{bommasani2021opportunities} significantly beneficial in classification tasks. They play a central role not only in
machine learning but also in a wide range of scientific domains, such as ChatGPT \cite{brown2020language,floridi2020gpt} for natural language, BloombergGPT \cite{wu2023bloomberggpt} for finance, DNABERT \cite{zhou2024dnabert,zhou2023dnabert,ji2021dnabert} for genomics, and many others \cite{zhou2024optimizing,qin2021ibert,Luo_2023,liu2022sciannotate,luo2024decoupled,yu2024enhancing,zhang2024smutf}. The Modern Hatfield Network provides another efficient method to do the question and answering classfication with the memory retrieval.
Modern Hopfield models \cite{hu2024nonparametric,hu2024computational,wu2023stanhop,hu2023SparseHopfield,hopfeildblog2021,ramsauer2020hopfield,hu2024outlier} showcase fast convergence and exponential memory capacity, linking them to Transformer architecture as advanced attention mechanisms. Their application spans various domains, including drug discovery \cite{schimunek2023contextenriched}, immunology \cite{widrich2020modern}, and time series forecasting \cite{wu2023stanhop,hu2023SparseHopfield,auer2023conformal}, signifying their influence on future large-scale model designs. 

\subsection{Chain-of-thought}
Another one method to do the classfication is chain-of-thought \cite{wang2022self,pan2024chainofaction,pan2024convcoa}. It enables the model to break down complex queries into a series of intermediate reasoning steps. This approach helps the model understand the context and nuances of the question, leading to more accurate and coherent answers. By sequentially processing each step, CoT enhances the interpretability and robustness of the QA system, ensuring that the final classification is based on a thorough and logical analysis of the input query.

\section{Model}
The task of QA classification takes the sentence S as an input, which varies the length sequence of the word $U=\{u_1, u_2, u_3, …, u_N\}$. For each word  $u_1 \in U$, there has a length value of $l_i \in L$ and a corresponding target label $y_i \in Y$, which represents the QA’s result associated with the corresponding sentence.
Figure 1 shows the overall architecture of the model, which involves several main components. (1) A self-attention algorithm to encode the sentence with the self-attention, (2) A Combination-level RNN to handle the output of the encoding and to classify the label of the sentence. We describe the details below.

\begin{figure}[h]
  \centering
  \includegraphics[width=\linewidth]{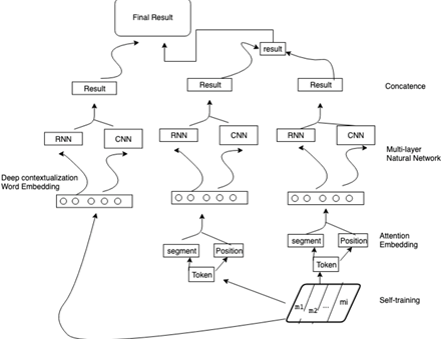}
  \caption{The graph of the model Architecture.}
\end{figure}
\subsection{Context-aware Self-learning}
Our self-attention algorithm encodes a variable-length sentence into a fixed size. There are two types of the algorithm; one based on Self–Attention and another based on deep contextualization word representation.

\subsubsection{Deep contextualization word representation}
This model uses the BiLM to consider the different position of words within the sequence. Inspired by Peters et al. \cite{peters2018deep}, we use PCA and t-SNE to reduce the dimensions from a higher level to reduce the dimensions from a higher level to a lower level. Then, we use the Combination-Level RNN (Section 3.2) which provides us with the previous hidden state of the encoded word. It provides us the contextual relationship in the sentences and combines all hidden states of words in sentences. After that, the deep modifications contextualization word representation encoder encodes the combination into the 2-D vectors of each sentence. We follow the instruction of Peters at el. \cite{peters2018deep} to explain below.	

A word $t_i$, which is the sequence of the sentence, is mapped onto the embedded layer. The deep contextualization representation uses BiLM to combine the forward and backend LM. The formulation of the process is as follows:
\begin{align*}
   \sum_{k=1}^N \left( \log p(t_k | t_1, \ldots, t_{k-1}; \Theta_x, \overrightarrow{\Theta}_{\text{LSTM}}, \Theta_s) + \log p(t_k | t_{k+1}, \ldots, t_N; \Theta_x, \overleftarrow{\Theta}_{\text{LSTM}}, \Theta_s) \right) 
\end{align*}
Moreover, we weigh the performance of the model with computing as indicated here:

$E(R_k;\Theta^{task}) = \gamma^{task}\sum^L_{j=0}s^{task}_jh^{LM}_{kj}$

In (1), the $s_j^{task}$ represents  softmax-normalized weights, and the scalar parameter $\gamma^{task}$ allows the task model to scale the entire vector. In the simple case, the representation would choose the top layer and  $E(R_k)$ = $h^{LM}_{kj}$.

\subsubsection{Self-Attention}
For each word in the word, we would use some Self-Attention model to encode them. The most popular Self-Attention model base is BERT \cite{devlin2018bert}. The model will encode a variable-length sequence using an attention mechanism that considers the different position token and segment within the sequence. Inspired by Devin et al. \cite{devlin2018bert} and Tran et al. \cite{tran2017hierarchical}, we apply the Combination-Level RNN (Section 3.2) into a self-attractive encoder \cite{lin2017structured}. We use the 24 layers and 1024 Hidden Uncased BERT also with the RobertaBERT as the base of the embedding to encode the context to the 3D tensor. We follow the instruction of Vipuls Raheja and Joel Tetreault \cite{raheja2019dialogue} and Joel Tetreault and Liu et al. \cite{liu2019roberta} to explain the modification mentioned below.

The word $t_i$ is also mapped onto the embedding layer and results in s-dimensional embedding for each word in the sequence based on the Transformer \cite{vaswani2017attention}. Then, the embedding is put into the bidirectional-GRU layer.

Vipul Raheja and Joel Tetreault \cite{raheja2019dialogue} describe the contextual self-attention score as:

    $S_i= W_{s2}\tanh(W_{s1}H^T_i + W_{s3}\overrightarrow{g_{i-1}}+b)$

Here $W_{s1}$  is a weight matrix, $W_{s2}$ and $W_{s3}$ is a matrix of parameters. b is a bias of the vector represented in Equation 2. This can be treated as a 2-layer MLP with bias, and $d_a$ with a hidden unit.

\subsection{Combination-level RNN}
The word representation $h_i$ from the past two models are passed into the combination-level RNN. Based on Figure 1, we would pass all of the hidden layers concatenated into a final representation $R_i$ of each word. The previous model step will help us build a encoding vector to represent the relationship of the each words, and in order to consider all the words in the sentence together to make the classification, it is necessary to make the combination RNN model.

This is more suitable for the problem classification to put the layers with the proper percentages in the final representation. Then, we place the result into the combination model layer to figure out the relationship between the label and the context of the words. This method is not independently decoding the label of the words; it should consider all of the relationships of the sentences. Then, it should determine the most related decoder to decode them to the related labels. The combination-level RNN would also have the function to supervise the labels and improve the accuracy of the classification of our model.

\subsection{Super-attractive}
The model that we use combines the final representative of the combination for hidden layers via self-attention. It can help us figure out what the labels of those words are and produce the results. The score we compute for the algorithm is to calculate the accuracy of the correct labels in the classifications as Hossin M. and Sulaiman M.N. \cite{hossin2015review} suggest. Also, we apply an advanced check for the question and answer problem. For sentences without clear results, we put them into the parser tree for another classification. The parser tree we use is based on Huang \cite{huang2019attentive}. We use its Tensor Product Representation to rebuild our parser tree for our model. The original Stanford Parser Tree  \cite{de2008stanford} is good to classify the relationship of the sentences. However, in our model, we use the Bi-LSTM with the attention algorithm to rebuild the parser tree and get the tree graph with POS tags. This is useful to calcify the structure of the sentence. After that, we use the graph we obtain to analyze the structure of words and produce the classification of the unsure sentence in the document. Finally, we determine the combination result for the users to check the question and answer problems.

\section{Data}
We evaluate the accuracy of the classification model with one standard dataset - the Switchboard Dialogue Act Corpus (SwDA) \cite{jurafsky1997switchboard} consisting of 43 classes (listed as , and we make a program to create the sentences based on the data with the Stanford Question Answering Dataset (SQuAD) \cite{rajpurkar2016squad} to use self-attention for the task. The Natural Language Toolkit Dataset (NLTK) \cite{loper2002nltk} is another significant resource for the test case. We use the \texttt{nltk.corpus.nps\_chat} dataset as data for the experiment. We then use the training, validation, and test splits as defined in Lee and Dernoncourt \cite{lee2016sequential}.

Table 1 shows the statistics for both datasets. There are many kinds of labels of the class to classify the kind of sentences they are. There are some special DA classes in both datasets, such as Tag-Question in SwDA and Statement-Question in NLTK. Both datasets make over  $25\%$ of the question type labels in each set.

\begin{table}
\centering
  \caption{Number of Sentences in the Dataset}
  \label{tab:freq}
  \begin{tabular}{c c c c c c}
    \toprule
    Dataset & Train & Validation & Test & T & N\\  [0.5ex]
    \midrule
    SwDA+SQuAD & 87k&10k&3k&43&100k\\
    NLTK &8.7k	&1k	&0.3k	&15	&10k\\
    \bottomrule
\end{tabular}
\end{table}

$|T|$ represents the number of classes and $|N|$ represents the sentence size

\section{Result}
We have compared the classification accuracy of our model(see Appendix) with several other models (Table 2). For methods using attention and deep contextualization word representation in some approaches to model the sentence of questionnaire documents, some of them use the self- attention word representation for the task. However, they did not perform as well as our model. All models and their variables were trained eight times, making an average for the performance as a result. And we find these previous algorithms did not perform as well as our model. Our model is better than Raheja and Tetreault \cite{raheja2019dialogue} by 0.4\% in SwDA dataset after measuring its accuracy score and 3.9\% for the Li and Wu \cite{li2019multi} methods in SWQA dataset. It also beats the TF-IDF GloVe baseline by 17.2\% in SwDA.

The improvements based on our model have a significant meaning for other models. However, the performance in NLTK is still not as good as that of the Raheja and Tetreault\cite{raheja2019dialogue} model. The reason for the lower accuracy is dependent on the contextual details and label noisy inside the dataset. The label noisy is caused with reason of that the NLTK dataset has the fewer and difference dialog act class, it would cause our model could not actually defined them by our model. The context in the NLTK dataset indicated the existence of some data not easily readable for the machine such as some error codes. Also, the label type in the NLTK dataset is only 35\% of the label type for the SwDA ones. As a result, due to the label noisy and the contextual details, the performance of NLTK did not show significant gains over that of SwDA.

\begin{table}
\centering
  \caption{QA Classification Accuracy of the different approaches}
  \label{tab:freq}
  \resizebox{0.65\textwidth}{!}{
  \begin{tabular}{ccl}
    \toprule
   Model & SwDA+SQuAD & NLTK \\
    \midrule
    TF-IDF GloVe \cite{pennington2014glove}&	66.1&	70.3\\
   Li and Wu \cite{li2019multi}&	79.2&	-\\
   Peters et al. \cite{peters2018deep}&	76.3&	- \\
   Raheja and Tetreault \cite{raheja2019dialogue}&	82.7&	85.8\\
   Lee and Dernoncourt  \cite{lee2016sequential}&	75.9&	77.4\\
   Our Method	&83.1&	85.5\\
   RoBERTa\cite{liu2019roberta}	&82.2&	84.7\\
  \bottomrule
\end{tabular}
}
\end{table}
The performance of our model is more sensitive than the model used commonly for the problems, including the error code. However, it has a higher accuracy, considering the complete problem classification. In future research, we should improve our algorithm, which has a higher ability to handle the problem of the label noisy and context detail.

\section{Conclusion}
We developed a new model which carefully performed the QA classification and made comparisons with common-use algorithms by testing the SwDA dataset. We used different word representation methods and determined that the context details depend highly on the classification performance. For example, the reason  NLTK is not as good as the Raheja and Tetreault \cite{raheja2019dialogue} results was because there were too many label noises and the context details which were not so easy to read. Working with attention and combination levels during the classification, which has not been previously applied in this kind of task, enables the model to learn more from the context and get more real meaning from the words than previously It helps to improve the performance of the classification for these kinds of tasks.

In our future work, we will explore more attention mechanisms, such as block self-attention \cite{shen2018bi}, or hierarchical attention \cite{yang2016hierarchical} and hypergraph attention \cite{bai2019hypergraph}. These approaches can incorporate the information from different representations for the various positions and can capture both local and long-range context dependency. Also, this approach should help with the problem of the hard-readable context, such as the problem of the NLTK dataset that causes accuracy to become lower than usual. We will seek more dataset combinations to do the question and answer classification work. We will also use RACE \cite{lai2017race} and GLUE  \cite{wang2018glue} datasets to do more test work and make more stable algorithms to solve the question and answer classification issues.

\section{Appendices}
\begin{table}[htp]
\centering
  \caption{Fine-tuning Hyperparameters of our model for each data set}
  \label{tab:freq}
  \begin{tabular}{cl}
    \toprule
    Hyperparam & SQuAD\\
    \midrule
    Learning Rate & 1e-5\\
    Weight Decay &	0.1\\
    Epochs&	7\\
    Batch Size&	8k\\
    
  \bottomrule
\end{tabular}
\end{table}

\begin{table}[htp]
\centering
  \caption{Pretraining Hyperparameters of our model}
  \label{tab:freq}
  \begin{tabular}{ccl}
    \toprule
    Hyperparam & RoBERTa& BERT\\
    \midrule
    No. of Layers &	24&	24\\
    Hidden Size &	1024&	1024\\
    FNN Inner Hidden &	4096&	-\\
    Attention Heads	&16	&16\\
    Attention Head size	&64	&64\\
    Dropout &	0.1	 &0.1\\
    Batch Size &	8k &	8k\\
    
  \bottomrule
\end{tabular}
\end{table}

\begin{figure}[htp]
  \centering
  \includegraphics[width=\linewidth]{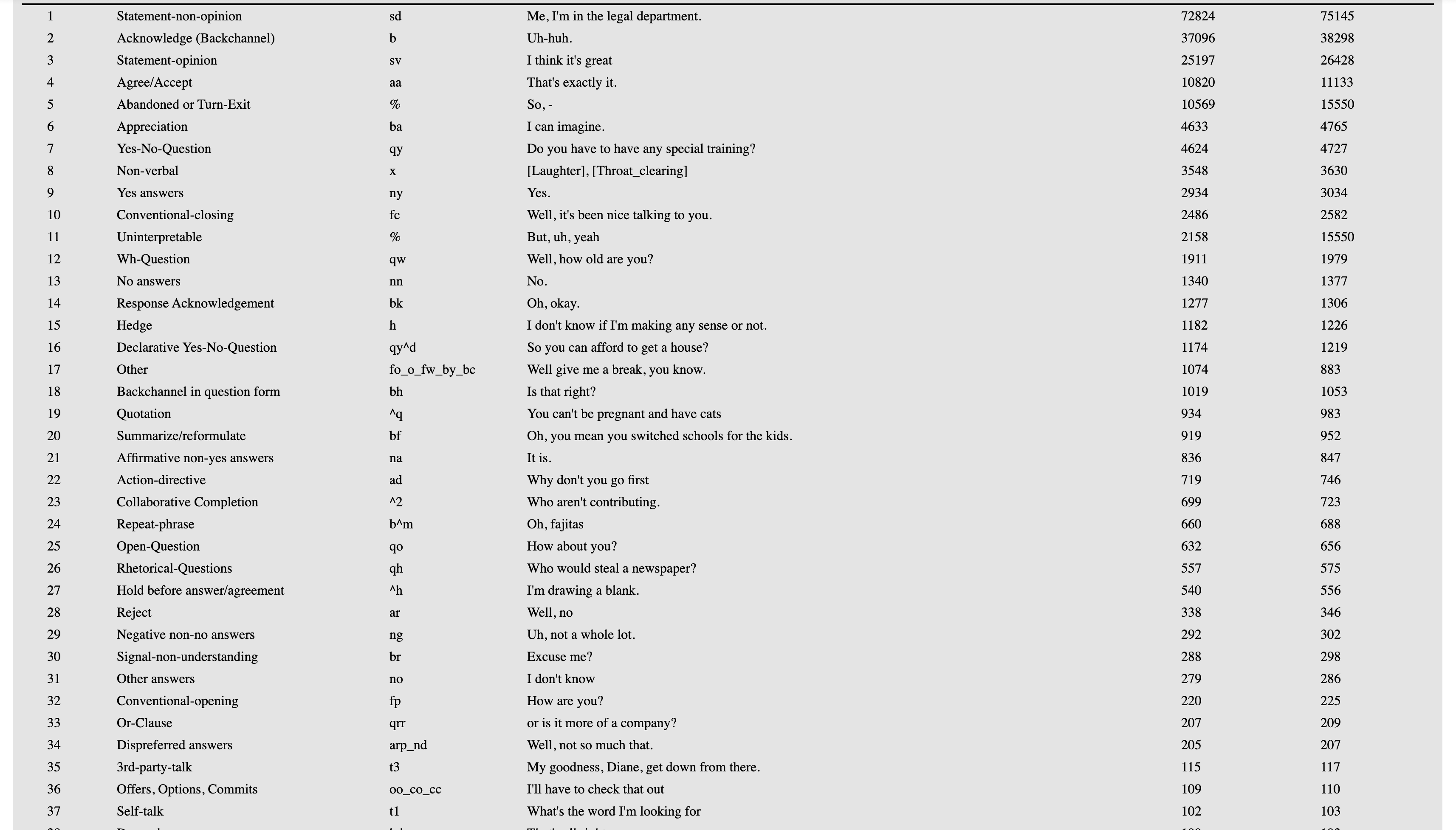}
  \caption{list of SwDA Dialog Act class and example\cite{jurafsky1997switchboard}}
\end{figure}

%
%
%
\newpage
\bibliographystyle{splncs04}
\bibliography{ref}

\end{document}